\documentclass[letterpaper, 10 pt, conference]{ieeeconf}  
\usepackage{graphics} 
\usepackage{epsfig} 
\usepackage{mathptmx} 
\usepackage{times} 
\usepackage{amsmath} 
\usepackage{amssymb}  
\usepackage{bm}
\usepackage{bigdelim}
\usepackage{setspace}
\usepackage{multicol, blindtext}
\usepackage{multirow}
\usepackage{url}
\usepackage{changes} 
\usepackage{amsmath}
\usepackage{pdflscape}
\usepackage{booktabs}
\usepackage{adjustbox}
\usepackage{pgfplots}
\usepackage{gensymb}

\usepackage{graphicx} 
\usepackage{subcaption} 
\usepackage{wrapfig}
\usepackage[shrink=50, selected=True]{microtype}

\newcommand{\HEADER}[1]{\ALC@it\underline{\textsc{#1}}\begin{ALC@g}}
	\newcommand{\ENDHEADER}{\end{ALC@g}}
\makeatother
\usepackage[pagebackref=true,breaklinks=true,colorlinks,bookmarks=false]{hyperref}
\usepackage{etoolbox}
\usepackage{longtable}%
\AtBeginEnvironment{longtable}{%
  \addfontfeature{RawFeature=+tnum;-onum}
}

\newcommand{\ie}{\textit{i}.\textit{e}.,}
\newcommand{\eg}{\textit{e}.\textit{g}.,}
\newcommand{\etal}{\textit{et~al}.}
\newcommand{\wrt}{\textit{w}.\textit{r}.\textit{t}.}
\newcommand{\cf}{\textit{cf}.}

\newcommand\copyrighttext{%
  \footnotesize © Accepted to ICRA 2024. Personal use of this material is permitted. Permission must be obtained for all other uses, in any current or future media, including
reprinting/republishing this material for advertising or promotional purposes, creating new collective works, for resale or redistribution to servers or lists,
or reuse of any copyrighted component of this work in other works.
}
\newcommand\copyrightnotice{%
\begin{tikzpicture}[remember picture,overlay]
\node[anchor=north,yshift=-10pt] at (current page.north) {\fbox{\parbox{\dimexpr\textwidth-\fboxsep-\fboxrule\relax}{\copyrighttext}}};
\end{tikzpicture}%
}

\title{\LARGE \bf{A Safety-Adapted Loss for Pedestrian Detection in Automated Driving}}

\author{Maria Lyssenko$^{\ast \ddagger}$, Piyush Pimplikar$^{\ast}$, Maarten Bieshaar$^{\ast}$, Farzad Nozarian$^{\ast\ast}$, Rudolph Triebel$^{\dagger\S}$\\
	\small${}^{\ast}$ Robert Bosch GmbH, Corporate Research, Germany, {\tt \small firstname.lastname@de.bosch.com}\\ 
	\small${}^{\ddagger}$ Technical University of Munich, Germany, {\tt \small firstname.lastname@tum.de} \\
    \small${}^{\ast\ast}$ German Research Center for Artificial Intelligence (DFKI), Saarbrücken, Germany, {\tt \small farzad.nozarian@dfki.de}\\
    \small${}^{\dagger}$ German Aerospace Center (DLR), Wessling, Germany, {\tt \small rudolph.triebel@dlr.de}\\
    \small${}^{\S}$ Karlsruhe Institute of Technology, Germany, {\tt \small rudolph.triebel@kit.edu}\\
}    

\begin{document}
\maketitle
\copyrightnotice

\begin{abstract}
In safety-critical domains like automated driving (AD), errors by the object detector may endanger pedestrians and other vulnerable road users (VRU). As common evaluation metrics are not an adequate safety indicator, recent works employ approaches to identify safety-critical VRU and back-annotate the risk to the object detector. However, those approaches do not consider the safety factor in the deep neural network (DNN) training process. Thus, state-of-the-art DNN penalizes all misdetections equally irrespective of their criticality. Subsequently, to mitigate the occurrence of critical failure cases, \ie~false negatives, a safety-aware training strategy might be required to enhance the detection performance for critical pedestrians. In this paper, we propose a novel safety-aware loss variation that leverages the estimated per-pedestrian criticality scores during training. We exploit the reachability set-based time-to-collision ($TTC_{RSB}$) metric from the motion domain along with distance information to account for the worst-case threat quantifying the criticality. Our evaluation results using RetinaNet and FCOS on the nuScenes dataset demonstrate that training the models with our safety-aware loss function mitigates the misdetection of critical pedestrians without sacrificing performance for the general case, \ie~pedestrians outside the safety-critical zone.
\end{abstract}


\section{Introduction}
\label{sec:intro}
Whenever autonomous mobile robots or automated vehicles (AV) operate in dynamic and highly complex environments ensuring correct and reliable detection of vulnerable road users (VRU) becomes vital. In respect thereof, current training and evaluation approaches of state-of-the-art object detectors have been extensively studied as an enabling technology due to the great success in camera-based perception~\cite{FCOS2019}.
In contrast to non-safety critical computer vision applications (CV) like \eg~parking occupancy detection in car parks or counting people for waiting time analytics~\cite{web},
cases of failed detections (so-called \textit{false negatives}) in automated driving (AD) scenarios may lead to a hazardous outcome~\cite{philion2020learning}. Thus, it is of utmost importance to assure accurate perception capabilities \wrt~classification and localization performance to avoid collisions with VRU.

\begin{figure}[htb]
	\centering
	\includegraphics[width=\columnwidth]{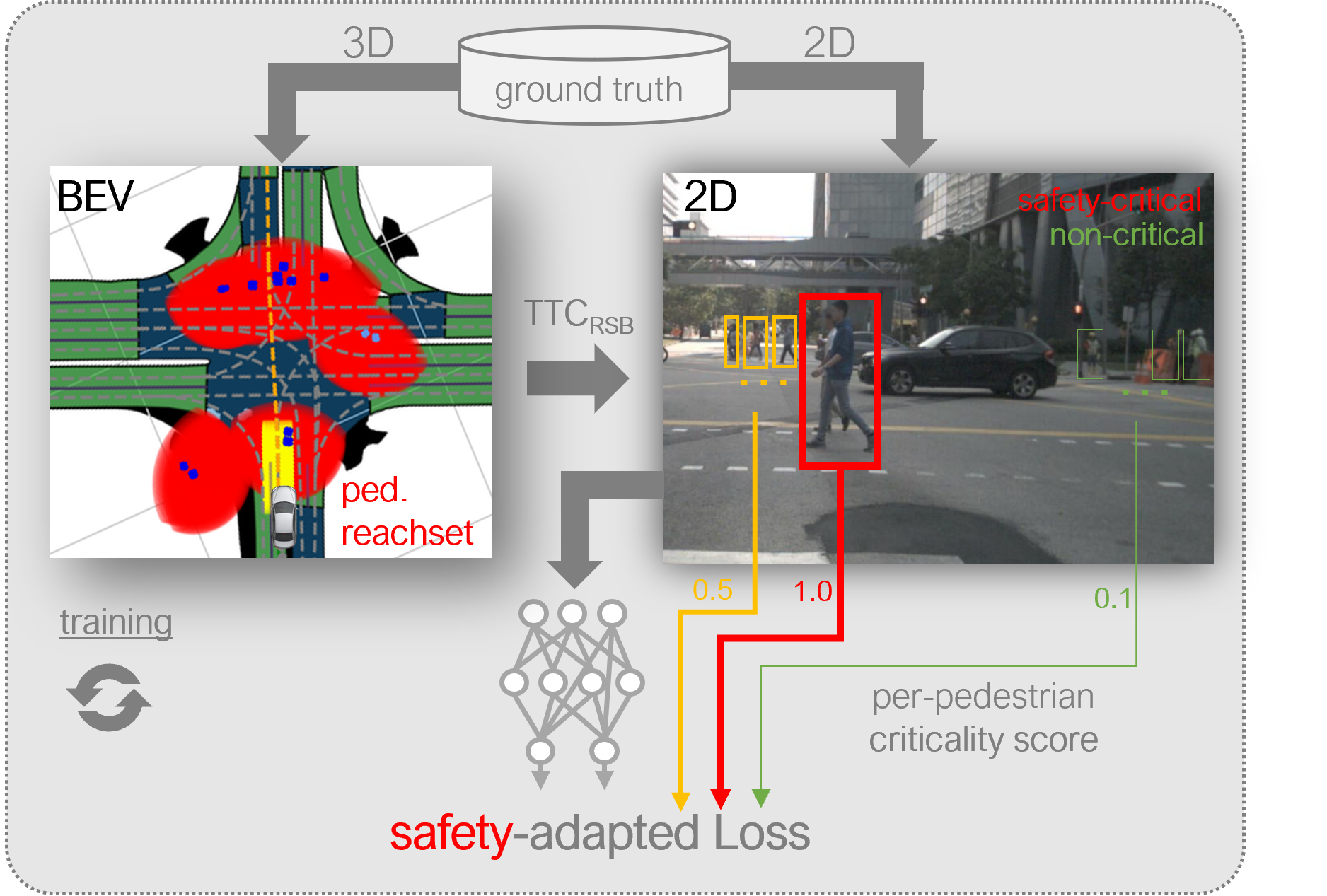}
	\caption{We propose the safety-adapted focal loss to enhance the detection performance of pedestrians under risk. The loss diligently exploits distance information and a threat metric from the motion domain (BEV) to consider the criticality of individual pedestrians during the training of an one-stage 2D object detector (vision domain). Concretely, we leverage reachable set-based time-to-collision (\textit{$TTC_{RSB}$}), defined as the intersection between the AV's (yellow) and pedestrians' (red) reachable sets, to quantify the \textit{per-pedestrian criticality} for each pedestrian in the scene. We back-annotate the \textit{criticality scores} of the detections to the \textit{safety-adapted loss} function and to dynamically adjust the loss contribution of safety-critical pedestrians.}
	\label{fig:eyecatcher}
\end{figure}

Let us consider a pedestrian detector applied to a crowded urban scene. 
Here, misdetections within the AVs safety zone pose an imminent collision risk (\eg~street crossing scenarios from Fig.~\ref{fig:eyecatcher} highlighted by the red bounding box) whereas distant misdetections do not yet affect the driving task (orange bounding boxes). 
Subsequently, the promotion of safe driving behavior necessitates \textit{(i)} the identification of all task-relevant pedestrians in the urban scene~\cite{Brkle2021SafePO,Mori2023ConservativeEO}, and \textit{(ii)} the guarantee of impeccable detection results for those under risk~\cite{Cheng2020SafetyAwareHO}\cite{Willers2020SafetyCA}.
Therefore, to tackle the underspecification of aggregated, purely vision-based metrics~\cite{DAmour2020UnderspecificationPC}, recent works by Wolf~\etal~\cite{wolf}, Bansal~\etal~\cite{bansal2021risk}, and Lyssenko~\etal~\cite{LGH+22} encompass a notion of criticality into the evaluation of the utilized object detector. 
As an example, in our investigated pedestrian detection use case in~\cite{LGH+22}, we employ the reachability set-based time-to-collision ($TTC_{RSB}$) determining the earliest point in time when a collision with the AV may occur.
Here, our evaluation has encountered several safety-critical misdetections within a significant number of identified sequences. 
Consequently, the question arises: \textit{How can we remedy potentially safety-critical pedestrian misdetections?}  

As our main contribution, we propose a novel, safety-refined loss function that effectively exploits the criticality of individual pedestrians during training as illustrated in Fig.~\ref{fig:eyecatcher}.
Therefore, we \textit{(i)} derive our per-pedestrian criticality score from the motion domain and
\textit{(ii)} we include the criticality into the focal loss to dynamically adjust the loss contribution \wrt~criticality.
Intuitively, by factoring in the criticality in the safety-adapted loss, we amplify the loss contribution of hazardous pedestrians and thus, focus the DNN on safety-critical cases. 
We provide an experimental evaluation of our safety-adapted loss utilizing the nuScenes dataset~\cite{nuscenes2019} and two state-of-the-art object detectors (RetinaNet~\cite{retina} and FCOS~\cite{FCOS2019}). 
Our results illustrate, that we successfully decrease pedestrian misdetections in the safety-critical zone with robust overall detection performance.

The remainder of the paper is structured as follows. First, we review the related work in Sec.~\ref{sec:relatedwork}, before introducing our methodology for constructing the safety-adapted loss function in Sec.~\ref{sec:method}. Thereafter, we provide the experiment setup in Sec.~\ref{sec:setup}, followed by the experimental results in Sec.~\ref{sec:results}.




\section{Related Work}
\label{sec:relatedwork}

\subsection{Task-Awareness in Automated Driving}
\label{sec:taskawareness}
Commonly used evaluation metrics like intersection-over-union (IoU), mean average precision (mAP), or recall are widespread because they are not specific to a particular task and allow for meaningful comparisons across different benchmarks~\cite{Zaidi2021ASO}. 
However, due to the task-agnostic character of those metrics there is no adequate assessment whether the perception function guarantees sufficient detection performance when deployed in the safety-critical AD domain~\cite{Willers2020SafetyCA}. 


The approaches by Wolf \etal~\cite{wolf}, Bansal \etal~\cite{bansal2021risk}, and Andrea \etal~\cite{Andrea2022} extend pure distance-based considerations~\cite{LGH+21,Gannamaneni2021SemanticCT}. They argue that a distance-based, \textit{potential} collision risk does not consider the dynamics and the criticality of the interaction. 
Therefore, the authors propose a definition of the \textit{imminent} collision risk implementing threat metrics from the AD domain such as TTC to account for a safety-indicator in aggregated evaluation measures.
However, the authors rely on simplified motion models assuming constant velocity vectors and heading charted over a time horizon.

Considering all possible, worst-case states an agent can reach within a time interval, recent works by Topan~\etal~\cite{Interactiondynamics,safetyzones} and Lyssenko \etal~\cite{LGH+22} implement the identification of task-relevant agents using dynamic-aware perception zones derived from reachability analyses.
Therefore, the work in \cite{Interactiondynamics} utilizes a Hamilton-Jacobi (HJ) reachability to construct a sound safety-zone around the AV whereas 
Lyssenko~\etal~\cite{LGH+22} assume that the AV is lane bound and implement a map-constrained calculation of the AV's reachable set leveraging motion models based on differential inclusions.
To account for uncertainties of future motion and the resulting worst-case criticality assessment, we utilize the framework proposed by Lyssenko~\etal\cite{LGH+22} to derive the $TTC_{RSB}$ for potentially dangerous interactions between pedestrians and AV.

\subsection{Significance of Loss Functions}
As one of the essences of the object detection task revolves around the significance of loss functions~\cite{autoped}. Therefore, recent progress has shifted from generic loss functions such as binary-cross-entropy (BCE) towards novel alternatives like the focal loss~\cite{retina,FCOS2019}, to reduce the importance of well-classified samples.

Further work by Li~\etal\cite{BoLi2022} extend the focal loss using a category-relevant, dynamic modulating factor to increase the impact of rare categories.
This also motivates our work to include a criticality component in the focal loss to magnify the loss contribution for individual pedestrians under risk.

To emphasize the 
detection capabilities \wrt~critical objects~\cite{Willers2020SafetyCA}, Cheng~\etal~\cite{Cheng2020SafetyAwareHO} published a conceptual proposal on safety specifications and mitigation strategies during the network construction process.
To the best of our knowledge, the first implementation of a safety-adapted regression loss is presented by Liao~\etal~\cite{Liao2023}. 
The authors additively combine the Smooth-$L_1$ with a safety loss component to minimize the discrepancy between prediction and ground truth for critical objects.
Therefore, the work proposes a safety criterion derived from the bird-eye-view (BEV) plane that quantifies the misalignment between the closest vertex \wrt~its distance.
However, for safety considerations, we primarily focus on the mitigation of misdetections (\ie~false negatives) and thus, propose a variant of the safety-adapted focal loss. 



\section{Methodology}
\label{sec:method}
In this section, we formulate our novel safety-adapted focal loss.
To mitigate the occurrence of dangerous misdetections, we construct our loss based on the dynamic-aware per-pedestrian criticality that encompasses the worst-case collision risk.
%
In Sec.~\ref{sec:reachsets}, we introduce the collision risk on the basis of $TTC_{RSB}$ from reachability analyses and present the combined per-pedestrian criticality in Sec.~\ref{sec:pedcrit}.
We motivate the design of our safety-adapted loss in Sec.~\ref{sec:safeloss}.


\subsection{Collision-Risk from Reachable Sets}
\label{sec:reachsets}

\begin{figure}[t]
	\centering
	\includegraphics[width=0.65\columnwidth]{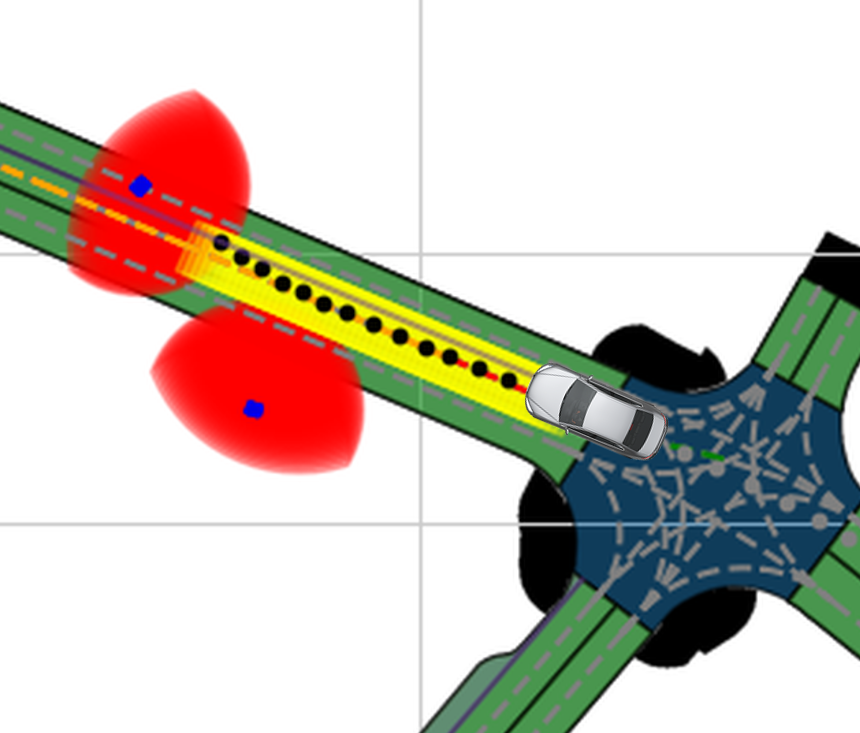}
	\caption{Exemplary illustration of a safety-critical interaction at time $\tau$ between an AV and pedestrians at risk (blue markers) and the respective reachable sets $R_{AV}(\tau)$ and $R_{ped,i}(\tau)$ for a given AV trajectory (black dots). The intersection of the reachable sets emphasizes the collision risk with $TTC_{RSB}\,<\,\infty$, \ie~$R_{AV}(\tau) \cap  R_{ped}(\tau) \neq \emptyset\ $. To calculate $R_{AV} (\tau)$ \wrt~the planned driving corridor~\cite{Schneider2021}, we require a sequence of the current (red) and successive centerlines (orange) as the input to the reachability framework from~\cite{LGH+22} to define our road.}
	\label{fig:reachset}
\end{figure}

The adequate assessment of an interaction's criticality between a pedestrian and the AV like in Fig.~\ref{fig:reachset}, requires a prediction on how the situation may evolve in the future, given the current state and the underlying motion models for the AV and the pedestrians, respectively.

In the following, we employ the reachability framework from previous work~\cite{LGH+22,Schneider2021} to estimate the per-pedestrian criticality from the AV's perspective.
Therefore, we exploit differential inclusion-based motion models to provide a safe overapproximation of the possible future states of the pedestrian and the AV, respectively~\cite{6784493}.

We leverage those models to acquire a clear specification of the objects' expected movement over a certain amount of time to calculate the so-called reachable set, \ie~a set of all possible future states the object could reach irrespective of the probability.
Thereby, we calculate the respective reachable sets from Fig.~\ref{fig:reachset} for each pedestrian $R_{ped,i}$ (red) and our automated vehicle $R_{AV}$ (yellow).
For the motion model definition, we utilize a constant acceleration model for the pedestrians and a constant velocity model from~\cite{LGH+22} for the AV.
In mathematical terms: For each pedestrian $i \in \lbrace 1,\ldots, N \rbrace$ in a scene with $N$ pedestrians, a given current system state $\bm{x}_i(\tau)$ at time $\tau \in \mathbb{R}^+$, its initialization for each $\bm{x}_i(0)$ (denoted by the initial position, velocity, and acceleration), and the constant acceleration motion model $F_a$ in terms of differential inclusions as $\dot{ \bm{x}_i}(\tau) \in F_a(\bm{x}_i(\tau))$, we can calculate the resulting reachable set 
\begin{gather}
R_{ped,i}(\tau)= \left \{\bm{x}_{i,u}(\tau) | \bm{x}_{i,u}(\tau)= \bm{x}_i(0) + \int_0^{\tau} u_i(t) dt \right \},\\
\text{where\;} \forall t_1 \leq \tau:  u_i(t_1)\in F_a(\bm{x}_i(t_1)).
\label{eq:abbrevInt2}
\end{gather}
Eq.~\ref{eq:abbrevInt2} means that $R_{ped,i}(\tau)$ contains all states $\bm{x}_{i,u}(\tau)$ that start at $\bm{x}_i(0)$ and can be reached by the trajectories $u_i(t)$ that are contained by $F_a(\bm{x}_i(t))$ with $t\in[0,\tau]$.
Please note, we perform the $R_{AV}$ calculation in accordance to Eq.~\ref{eq:abbrevInt2} employing the constant velocity model $F_v(\bm{x}_{AV}(t))$.

To estimate the criticality between each pedestrian $i$ and the AV, we use time-to-collision (TTC) as our threat metric, which quantifies the earliest point in time
\begin{gather}
TTC_{RSB,i} = \min\{\tau \mid  R_{AV}(\tau) \cap  R_{ped,i}(\tau) \neq \emptyset\}
\end{gather}
when two reachable sets intersect, \ie~the first point in time where a collision \textit{may} happen.
Following Schneider~\etal~\cite{Schneider2021}, we exploit a TTC formulation based on reachable sets to extend the current state-of-the-art TTC formulation that utilizes point estimates without uncertainties~\cite{DROF19}.
Please note, there are current deep learning approaches to predict the TTC from mono-camera input utilizing, \eg~optical flow~\cite{binaryTTC}. However, we opt to use precise ground truth information to facilitate white-box methods for safety-considerations~\cite{Willers2020SafetyCA}.  

\subsection{Per-Pedestrian Criticality}
\label{sec:pedcrit}
The initial work on $TTC_{RSB,i}$ in~\cite{LGH+22} exploits reachable sets to identify pedestrians at risk of an imminent collision below a $TTC$ threshold, \ie~$TTC_{RSB,i}\,<\,TTC_{crit}$.
However, the study revealed that the employment of the constant velocity model for $R_{AV}$ calculation may produce a \textit{blind spot}, \ie~an insensitivity for $TTC_{RSB,i}\,>\,TTC_{crit}$.

Concretely, given a low AV velocity, the reachability analyses produces a $R_{AV}$ of a small spatial extent and thus, a non-critical $TTC_{RSB,i}$ although the pedestrian is in the direct vicinity of the AV.
Therefore, we 
additionally inject distance information to account for the potential collision risk for non-critical pedestrians~\wrt~$TTC_{RSB,i}$.

\begin{wrapfigure}{R}{0.15\textwidth}
\centering
\begin{tikzpicture}
      \draw[->, line width=0.3mm] (-0.25, 0) -- (2, 0) node[right] {$d_i$};
      \draw[->, line width=0.3mm] (0, 0) -- (0, 2) node[above] {$\kappa_{d,i}$};
      \draw[scale=1.5, line width=0.3mm, domain=0:1, smooth, variable=\x, blue] plot ({\x}, {-\x*\x+1});
      \draw[-, line width=0.3mm] (-0.15,1.5) -- (0.15,1.5);
    \draw[-, line width=0.3mm] (1.5,-0.15) -- (1.5,0.15);
      \node[] (d) at (1.5,-0.35) {$d_{\textit{max}}$};
      \node[anchor=north east] (c) at (0,1.9) {$1$};
      \node[] (d) at (-0.2,-0.2) {$0$};
    \end{tikzpicture}
\caption{\label{fig:distcrit}Downward parabola to model the distance criticality $\kappa_{d,i}$ over the distance $d_i$ to the AV for a pedestrian $i$.}
\end{wrapfigure}
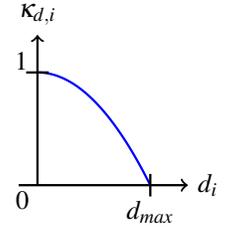

Based on this underlying idea, we compose our criticality weighting for individual pedestrians $\kappa_i$ by means of \textit{(i)} $TTC_{RSB,i}$ that accounts for the collision criticality ($\kappa_{c,i}$) considering the uncertainty-aware dynamics of the interaction, and \textit{(ii)} the distance between the pedestrian and AV to reflect the distance criticality ($\kappa_{d,i}$) irrespective of the motion model.

Let us now focus on the implementation of $\kappa_i(\kappa_{c,i},\kappa_{d,i})$, where we want to design $\kappa_i\in[0,1]$, \ie~$\kappa_i=1$ represents a pedestrian of highest relevance for the driving task.
Inspired by the work of Ceccarelli~\etal~\cite{Andrea2022}, we utilize the downward parabola from Fig.~\ref{fig:distcrit} that passes through the points $(0,1)$ and $(d_{max},0)$, to describe $\kappa_{d,i}$ over distance $d_i$. 
Please note, $d_{max}$ describes the distance up to which we consider a pedestrian as safety-relevant for the driving task.
Further, we leverage the non-linear decrease of $\kappa_{d,i}$
\begin{gather}
    \kappa_{d,i}(d_i) = -\frac{1}{d_{max}^2}d_i^2+1,~\text{where}~d_i\in[0,d_{max}]\land \kappa_{d,i}\in[0,1]
\label{eq:distcrit}
\end{gather}
to achieve a slow decrease in the distance criticality for $d_i\xrightarrow~0$, \ie~for pedestrians close to the AV.
Thus, for distant pedestrians with $d_i\xrightarrow~d_{max}$, we estimate $\kappa_{d,i}\xrightarrow~0$.
To this end, we apply Eq.~\ref{eq:distcrit} for $\kappa_{c,i}\in[0,1]$, respectively, to estimate the collision criticality according to the non-linear decrease \wrt~the time $t$. 
Here, we use $TTC_{max}$ as the time threshold leading to $TTC_{RSB,i}\xrightarrow~TTC_{max}$ and $\kappa_{c,i}\xrightarrow~0$ for distant pedestrians.
For the composed per-pedestrian criticality $\kappa_i$~\cite{wolf}
\begin{gather}
\kappa_i = \frac{1}{3}(2\kappa_{c,i}+\kappa_{d,i}) \\ \text{where}~\kappa_{c,i},\,\kappa_{d_i},\,\kappa_i\in[0,1],
\end{gather}
we implement a double weighting of $\kappa_{c,i}$ in the formula as it includes a dynamics-aware criticality estimate and thus, a superior measure of the collision risk. 

\subsection{Safety-adapted Focal Loss}
\label{sec:safeloss}
The focal loss (\cf~\ Eq.~\ref{eq:FL}) is commonly employed in object detectors to mitigate the foreground-background imbalance~\cite{retina}.
Thereby, the key idea of the loss is to re-balance the loss contribution of easy samples, \ie~decrease their importance in the training process.
As described in~\cite{retina}, the focal loss \textit{FL}
\begin{equation}
FL(p_{i}) = -\alpha(1-p_{i})^{\gamma}\log(p_{i})
\label{eq:FL}
\end{equation}
implements an extension of the generic \textit{BCE} loss by utilizing a weighting factor $\alpha$ and a modulating factor $(1-p_i)^\gamma$. 
Given the definition, the focusing parameter $\gamma$ determines the properties of the \textit{FL} and down-weights easy samples \wrt~to the predicted object's class probability $p_i\in[0,1]$.
Concretely, a higher $\gamma$ extends the probability range when a sample is considered easy and thus, lowers the loss contribution for those well-classified samples accordingly.
However, the current loss determination is solely dependent on $p_i$ and the hyperparamers ($\alpha,\gamma$) irrespective of the objects' criticality. 
To tackle the importance of objects, Li~\etal~\cite{BoLi2022} emphasize a category-dependent focusing factor for a long-tail distribution, \ie~in our case the imbalance between critical and non-critical pedestrians (see Fig.~\ref{fig:pedcount}).

Inspired by Li~\etal~\cite{BoLi2022}, we want to inject the criticality into the loss but on an instance level, \ie~for individual pedestrians, as for safety considerations there should be a discrimination between task-relevant and task-irrelevant objects within a category.
In our work, we leverage the criticality $\kappa_i$ from Sec.~\ref{sec:pedcrit} to amplify the loss contribution for critical pedestrians with $\kappa\xrightarrow~1$ in our safety-adapted \textit{FL}
\begin{equation}
FL_{\kappa_i}(p_{i}, \kappa_i) = -\alpha(1-p_i)^{(\gamma-\kappa_i)}log(p_i).
\end{equation}

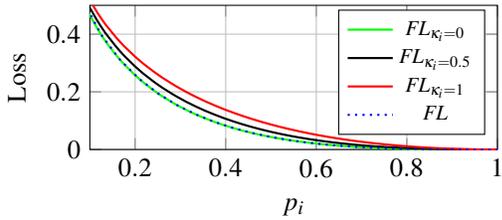
\begin{figure}[t]
    \centering
    \begin{tikzpicture}
      \begin{axis}[
        xlabel=$p_i$,
        ylabel= Loss,
        xmin=0.1, xmax=1,
        ymin=0, ymax=0.5,
        grid=major,
         ylabel style={
      yshift=-7.5pt 
    },
        width=7cm,
        height=3.5cm,
        legend pos=north east,
        legend style={font=\footnotesize}
      ]
      \addplot[domain=0:1,smooth,thick,green] {-0.25*(1-x)^2 * ln(x)};
       \addplot[domain=0:1,smooth,thick,black] {-0.25*(1-x)^(1.5) * ln(x)};
      \addplot[domain=0:1,smooth,thick,red] {-0.25*(1-x)^1 * ln(x)};
     \addplot[domain=0:1,dotted,thick,blue] {-0.25*(1-x)^2 * ln(x)};

      \addlegendentry{$\textit{FL}_{\kappa_i=0}$}
        \addlegendentry{$\textit{FL}_{\kappa_i=0.5}$}
       \addlegendentry{$\textit{FL}_{\kappa_i=1}$}
        \addlegendentry{$\textit{FL}$}
      \end{axis}
    \end{tikzpicture}
    \caption{Illustration of the focal loss \textit{FL} (with $\gamma=2, \alpha=0.25$) and the safety-adapted focal loss $FL_{\kappa,i}$ \wrt~the pedestrians' criticality $\kappa_i$. For $\kappa_{i}=0$ (non-critical pedestrians), we preserve the properties of the \textit{FL}, \ie~$FL_{\kappa,i}\xrightarrow~FL$. For critical pedestrians with $\kappa\xrightarrow~1$, we naturally magnify the loss contribution, where $FL_{\kappa\xrightarrow~1}\geq~FL$.}
    \label{fig:focalloss}
\end{figure}

Given the properties of the \textit{FL}, a larger $\gamma\,>2$ is used for a severe positive-negative imbalance, which will result in a sacrifice \wrt~to the samples' loss contribution in the training process.
This limits the performance for rare samples, \ie~safety-critical pedestrians. 
With our instance-based criticality weighting $\kappa_i$, we want to counteract the diminishing loss contribution for critical pedestrians and propose the adaptation of the focusing parameter to $(\gamma\textit{-}\kappa_i)$ with $\gamma=2$~\cite{retina} as illustrated in Fig.~\ref{fig:focalloss}.

Thereby, we \textit{(i)} dynamically change the loss contribution for $\kappa_i>0$ \wrt~a pedestrian's criticality in the training process, and \textit{(ii)} we maintain the properties of the \textit{FL} for non-critical pedestrians, \ie~with $\kappa_i\xrightarrow~0$ we obtain $\textit{FL}_{\kappa_i}\xrightarrow~\textit{FL}$.
Please note, for our weighting we derive $\kappa_i$ directly from the motion domain, \ie~no extensive hyperparameter search is required.

\section{Experiment Setup}
\label{sec:setup}
In the following, we describe the setup to evaluate our novel safety-adapted loss. In Sec.~\ref{sec:datasets} and Sec.~\ref{sec:datacuration}, we introduce the utilized datasets, and the data cleaning procedure, respectively.
In Sec.~\ref{sec:protocol}, we elaborate on the training protocols for the employed pedestrian detectors.

\subsection{Datasets from nuTonomy}
\label{sec:datasets}
For our experiments, we employ two datasets from nuTonomy: nuImages and nuScenes~\cite{nuscenes2019}.
We use the 2D nuImages dataset with its precise bounding box annotations for the pedestrian object detector's initial pre-training.

As the criticality estimation necessitates domain information, we require a 3D dataset with position and velocity information of the pedestrians along with map information for the AV to calculate the $TTC_{RSB,i}$ and distance information, respectively.
Therefore, we utilize nuScenes as it provides data from the entire sensor suite of an AV for 1000 scenes. We perform the training and evaluation on the splits as defined in the \textit{nuScenes-devkit}~\cite{nuscapi}.
\begin{wrapfigure}{R}{0.2\textwidth}
    \centering
    \includegraphics[width=0.3\columnwidth]{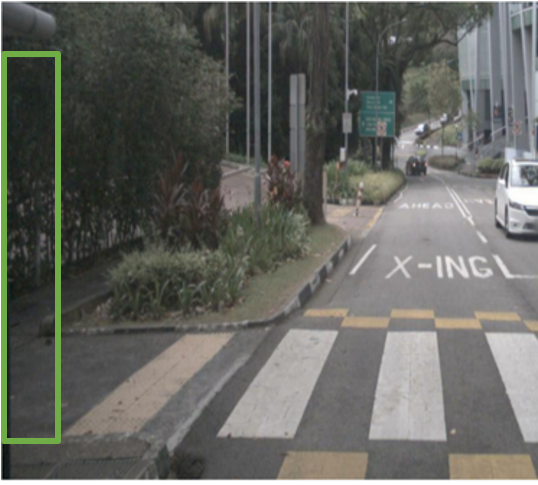}
    \caption{\label{fig:misleading} Invalid 2D bounding box due to nuScenes' annotation policy.}
\end{wrapfigure}

Please note, for our experiments, we utilize the images from the front camera only, and the corresponding lidar point clouds of the scene that are matched over a scene \texttt{token}.
However, despite the rich annotations, the nuScenes dataset contains only 3D bounding boxes.
Therefore, we project the cuboids' coordinates onto the camera pixel grid to retrieve 2D pedestrian annotations using the helper function \texttt{get\_2D\_boxes()}.

\subsection{Data Curation on nuScenes}
\label{sec:datacuration}
The nuScenes labelling policy discards object boxes without any lidar and radar points to filter out temporarily fully occluded objects~\cite{nuscerr}.
However, there are still occurrences of false positive annotations as illustrated in Fig.~\ref{fig:misleading}.
As the \texttt{get\_2D\_boxes()} function projects the cuboids into the frames of all cameras, bounding box projections from the left and right cameras might appear in the relevant front camera frame.
To mitigate such artifacts, we utilize the pedestrians' position information from the motion domain.
Thereby, for each projected box, we determine whether the center of its cuboid falls within the AV's physical field-of-view of the front camera, which is 70$\degree$.
In the case of a missing correspondence, we associate the cuboid with one of the side cameras and discard the respective 2D annotation from the front camera in the training and evaluation phase.



\subsection{Training Protocol: Pedestrian Detectors}
\label{sec:protocol}
For our pedestrian detectors, we implement the RetinaNet~\cite{retina} and the FCOS~\cite{FCOS2019} using PyTorch~\cite{pytorch} and employ the following pre-training protocols on nuImages.

\textbf{RetinaNet}: In the implementation from~\cite{retinayeh}, we utilize the ResNet-50 backbone, the Adam optimizer with a learning rate of $1e^{-5}$, the reduce-on-loss-plateau scheduler (patience=3), and we train our model for 200 epochs using a batch size of 16.
We obtain an AP50 of $0.31$ for the pedestrian class on the nuImages validation split.

\textbf{FCOS}: We follow the original paper implementation from~\cite{fcostian} with the ResNet-50 backbone that is trained for 42 epochs using the batch size of 16.
During training, we employ the stochastic gradient descent optimizer with an initial learning rate of $1e^{-3}$. Furthermore, we apply a multi-step learning rate decay with a linear warm-up.
Here, we obtain an AP50 of 0.48 for the pedestrian class.

\textbf{Safety-adapted training}: For the implementation of the safety-adapted loss from Sec.~\ref{sec:safeloss} for both pedestrian detectors, we use the respective models pre-trained on nuImages as they indicated a reasonable performance for the pedestrian class. 
With the safety-adapted loss, we train the models on the nuScenes' training split until the loss on the validation set converges ($\approx4$~epochs).
More specifically, we leverage the estimated criticality from Sec.~\ref{sec:pedcrit} to dynamically adapt the modulating factor for the pedestrian class.
For other categories like cars and the background class, we set $\kappa=0$ to maintain the the properties of the focal loss.

\section{Experimental Results}
\label{sec:results}
 In Sec.~\ref{sec:evalloss}, we present the evaluation of our novel safety-adapted focal loss for RetinaNet and FCOS \wrt~the focal loss baseline, and investigate the impact on detection capabilities for pedestrians of a different criticality.
 Further, in Sec.~\ref{sec:easyhard} we relate the safety-adapted loss to the pedestrians' detection easiness, and in Sec.~\ref{sec:ablation}, we analyze how the design of the per-pedestrian criticality affects the safety-critical performance.

\subsection{Safety-Adapted Loss Evaluation}
\label{sec:evalloss}
\begin{figure}[t]
	\centering
	\includegraphics[width=\columnwidth]{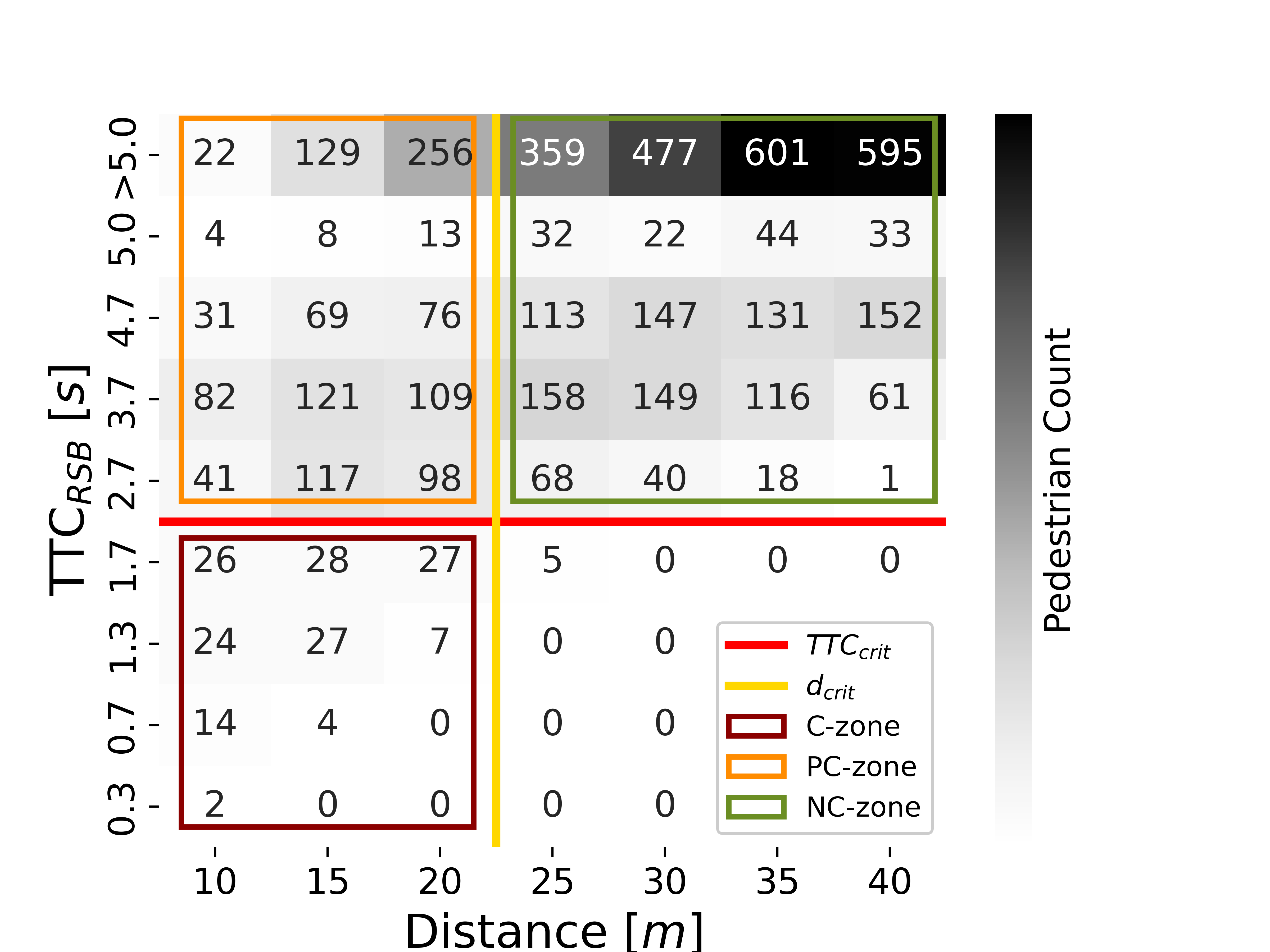}
	\caption{The heatmap illustrates the pedestrian count \wrt~the corresponding TTC$_{RSB}$ and distance values for the safety-critical $\bm{C}$ (159, \textit{dark red}), potentially critical $PC$ (1126, \textit{orange}) and non-critical (3317 for $d>20\,m$, \textit{green}) zone. The safety-critical zone is defined by TTC$_{crit}$ and dist$_{crit}$.}
	\label{fig:pedcount}
\end{figure}

We start our evaluation by defining three zones (with the corresponding pedestrian count) that encompass the critical (159), potentially critical (1126) and non-critical pedestrians (3371) as illustrated in the heatmap in Fig.~\ref{fig:pedcount}.
For a distance $>40\,m$ we have additional 3025 non-critical pedestrian instances that are not visualized in Fig.~\ref{fig:pedcount}.

Depending on the $TTC_{RSB}$ and the distance for individual pedestrians, each cell summarizes the respective count in the nuScenes validation set.
Please note, that the lower right part of the heatmap does not contain any samples as the AV's velocity is thresholded by the urban speed limit of up to 30~$mph$~($\approx13,3ms^{-1}$). Hence, the speed limit lower-bounds the $TTC_{RSB}$ that is feasible for a given distance.

Consequently, for the given speedlimit a braking time of 1.7$s$ might be required to avoid a collision~\cite{LGH+22}.
Therefore, we define the safety-critical zone $\bm{C}$ by \textit{(i)} $TTC_{crit}=1.7\,s$, and \textit{(ii)} a critical distance of dist$_{crit}=$20$\,m$ that contains roughly $2.1\%$ of all pedestrians.
The potentially safety-critical zone $(PC)$ is lower-bounded by TTC{$_{crit}$} with a distance up to $d_{crit}$ with $14.8\%$ of all pedestrians, and the non-critical zone $(NC)$ contains the remaining pedestrians of the validation set.
Given the discussed blind spot for a low AV's velocity in Sec.~\ref{sec:pedcrit} (high $TTC_{RSB}$ below $d_{crit}$), we explicitly consider the potentially critical zone $PC$ in our evaluation as the AV still may accelerate to the speed limit and thus, shift the pedestrians into the safety-critical zone $\bm{C}$.

\begin{table}[t]
    \centering
    \caption{Performance of RetinaNet and FCOS on nuScenes utilizing the baseline focal loss $FL_{\gamma=2}$ and the safety-adapted focal loss $FL_{\kappa}$. AP$^{S}$, AP$^{M}$, and AP$^{L}$ show AP$^{50}$ scores for \textit{all} pedestrians conditioned on the bounding box diagonals, \ie~$<150\,px~\text{(small)}, 151\,\text{-}350\,px~\text{(mid)}, \text{and} >350\,px~\text{(large)}$. Recall$^{\bm{C}}$ and Recall$^{NC}$ define the recall for critical and non-critical pedestrians.}
    \label{tab:overallperf}
    
    \begin{adjustbox}{width=\columnwidth}
        \begin{tabular}{lccccccccc}
            \toprule

             Model & Method & AP$^{50}$ & AP$^S$ & AP$^M$ & AP$^L$ & Recall$^{\bm{C} }$ & Recall$^{PC}$ & Recall$^{NC}$ & Precision  \\
            \midrule
             & ${FL_{\gamma=2}}$ & 0.441 & 0.366 & 0.667 & 0.656 & 0.881 & 0.790 & 0.441 & 0.782 \\
            RetinaNet$-$50 & ${FL_{{\gamma=1}}}$ & 0.429 & 0.347 & 0.691 & 0.746 & 0.899  & 0.775 & 0.411 & 0.830\\
             & ${\bm{FL}_{\bm{\kappa}}}$ & 0.440 & 0.356 & \textbf{0.729} & 0.656 & \textbf{0.906}  & \textbf{0.803} & 0.433 & 0.778\\
            \hline
              & ${FL_{{\gamma=2}}}$ & 0.476 & 0.424 & 0.616 & 0.615 & 0.918  & 0.879 & 0.649 & 0.316 \\
             FCOS$-$50 &${FL_{\gamma=1}}$ & 0.457 & 0.412 & 0.667 & 0.521 & 0.830  & 0.796 & 0.535 & 0.320\\
              & ${\bm{FL}_{\bm{\kappa}}}$ & 0.474 & 0.422 & 0.663 & 0.594 & \textbf{0.950} & 0.878 & 0.644& \textbf{0.322}\\
            \bottomrule
        \end{tabular}
    \end{adjustbox}
\end{table}

In Tab.~\ref{tab:overallperf}, we outline the effectiveness of our approach for the safety-critical zone $\bm{C}$.
Therefore, we compare the safety-adapted loss $FL_\kappa$ to the baseline focal losses ${FL_{\gamma=1}}$ and ${FL_{\gamma=2}}$.
The table demonstrates that in comparison to $FL_\kappa$, the "vanilla" reduction of the focusing parameter to $\gamma=1$ decreases the recall scores (Recall$^{\bm{C}}$) for RetinaNet and FCOS by 0.7$\%$ and 12$\%$, respectively. 
Similarly for $PC$, we denote a performance decline for both models by 0.8$\%$ and 8.2$\%$ when evaluated with $FL_{\gamma=1}$. 
Thereby, from the results we can conclude that a naive decrease in $\gamma$ for all samples (and consequently a higher loss contribution for all samples irrespective the criticality) leads to a diminished recall in $\bm{C}$ and $PC$, \ie~a higher number of (potentially) critical misdetections.

Thus, the results reflect that the dynamic adjustment of the focusing parameter $(\gamma-\kappa)$ achieves superior sensitivity in $\bm{C}$ in comparison to ${FL_{\gamma=1}}$ and ${FL_{\gamma=2}}$.
Particularly for RetinaNet and FCOS in $\bm{C}$, our approach improves the ${FL_{\gamma=2}}$ recall baseline by 2.5$\%$ and 3.2$\%$, respectively.
For $PC$, our evaluation demonstrates a recall incline by 1.3$\%$ for the RetinaNet and stable performance for FCOS.

Conclusively, the results from Tab.~\ref{tab:overallperf} emphasize that {$FL_\kappa$} can handle the severe imbalance between critical, potentially critical, and non-critical pedestrians with a negligibly small performance degradation of $<1.0\%$ in the overall performance (\cf~AP$^{50}$, AP$^{S}$, AP$^{M}$, AP$^{L}$, and Precision in Tab.~\ref{tab:overallperf}).
Please note, that for our zone-based evaluation we primarily employ recall scores as we require physical properties to identify pedestrians as critical, potentially critical, or non-critical. 
False positive detections lack associated 3D ground truth. Hence, we can not calculate the precision for the three zones trivially.
We leave this for future work.
However, in the overall precision and AP$^{50}$ scores for RetinaNet, we observe only a minor decrease of up to $0.4\%$.
Specifically for FCOS, there is even a further improvement in precision by 0.6$\%$.


\begin{figure}[t]
	\centering
	\includegraphics[width=\columnwidth]{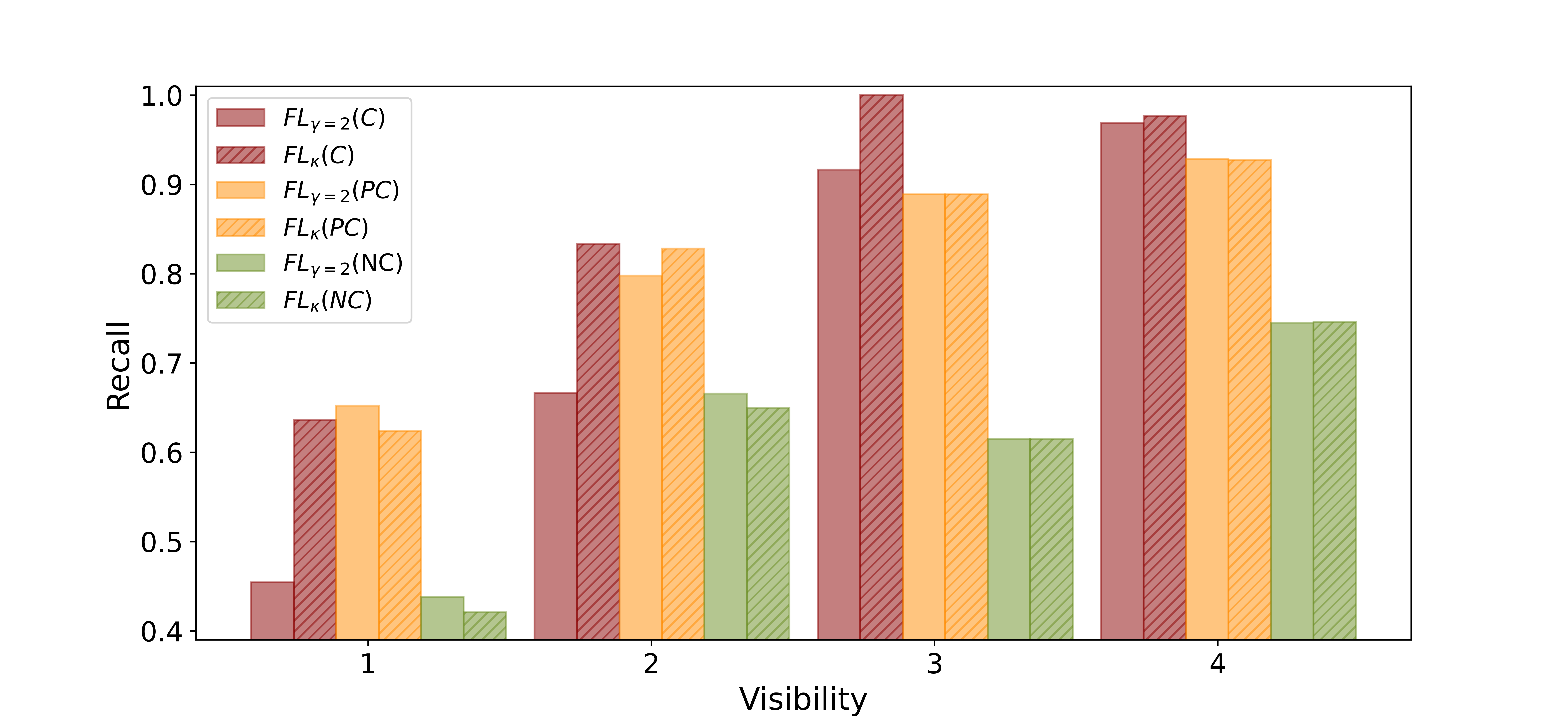}
	\caption{The bar plot depicts the recall performance of FCOS \wrt~the visibility from 1 (hardest) to 4 (easiest) for critical, potentially critical, and non-critical pedestrians evaluated with $FL_\kappa$ and $FL_{\gamma=2}$.}
	\label{fig:barplot}
\end{figure}

\subsection{Ablation Study: Criticality and Learning Difficulty}
\label{sec:easyhard}
In this ablation study, we investigate to which extent the safety-adaptation of the loss relates to the detection's easiness or hardness.
In other words, we want to ensure that our model trained with $FL_\kappa$ does not only learn to detect "easy" but critical pedestrians, \eg~unoccluded pedestrians in the direct vicinity those that are easily visible. Among numerous definitions~\cite{ZW23} to define detection and sample difficulty, we use the annotated visibility from nuScenes dataset as a simple proxy.

We conduct the detailed analyses on the FCOS as it outperformed RetinaNet by 3.4$\%$ and 4.4$\%$~\wrt~AP$^{50}$ and Recall$^{\bm{C}}$. 
The bar plot in Fig.~\ref{fig:barplot} depicts the recall values for the pedestrians within the zones of different criticality from Fig.~\ref{fig:pedcount}, \ie~for the zones $\bm{C},\, PC$ and $NC$, evaluated with $FL_\kappa$ and the $FL_{\gamma=2}$, respectively, for different partitions of pedestrian visibilities (4 bins representing a decreasing difficulty )~\cite{nuscenes2019} up to 40$\,m$.
As expected, for all categories the distribution shows an increasing recall trend with a higher visibility (easier samples).
The bar plot also illustrates approximately equal recall scores for potentially critical and non-critical pedestrians from $PC$ and $NC$ for higher visibilities.
Given our definition of $FL_\kappa$, we would expect that behaviour as we design $FL_\kappa\xrightarrow~FL_{\gamma=2}$ for pedestrians of lower criticality with $\kappa\xrightarrow~0$.
For lower visibilities (bin $1$ and $2$), the distribution shows that except for $PC$ our safety-adapted focal loss $F_\kappa$ outperforms the baseline $FL_{\gamma=2}$.
Particularly, for the critical zone, we denote a recall increase for all visibilities, which indicates that we are able to mitigate false negatives within partitions of different difficulty.

\subsection{Ablation Study: TTC vs. Distance}
\label{sec:ablation}
In our second ablation study, we evaluate the impact of individual components of the composed per-pedestrian criticality on the performance. 
Therefore, in Tbl.~\ref{tbl:critablation}, we decouple the distance criticality $\kappa_d$ and the collision criticality $\kappa_c$ from Sec.~\ref{sec:pedcrit} into individual losses ($FL_{\kappa_d}$ and $FL_{\kappa_c}$) and compare the results to the baseline $FL_\kappa$.
As in Sec.~\ref{sec:evalloss}, we perform the recall evaluation for our three zones of different criticality.

The results show that there is a different trend  for our evaluated losses \wrt~a particular criticality.
As we can see, ${FL_{{\kappa_c}}}$ outperforms the baseline $FL_\kappa$ among almost all categories for the RetinaNet.
It shows also superior results to the pure distance criticality $FL_{{\kappa_d}}$ except for AP$^{L}$ with a negligibly small decrease of only $0.7\%$.
However, for FCOS the performance scores do not reflect a sound indication for a particular criticality choice.
From the recall evaluation of the three zones, we can derive that for zone $\bm{C}$ pure distance information was \textit{(i)} sufficient to reach the performance from the baseline, and \textit{(ii)} able to achieve a higher recall for $PC$ by almost $1\%$, while maintaining stable results within $NC$.

In conclusion, although our $TTC_{RSB}$-based criticality from Sec.~\ref{sec:reachsets} promotes the identification of safety-critical pedestrian \wrt~dynamic properties, our ablation study has shown that depending on the model, a simple criticality measure like distance may be employed as a reasonable proxy for relevance during training.

\begin{table}[t]
    \centering
    \caption{Extension of Tbl.~\ref{tab:overallperf} to illustrate the ablation study on the composed criticality $\kappa$ \wrt~collision criticality $\kappa_c$ and distance criticality $\kappa_d$.}
    \label{tbl:critablation}
    \begin{adjustbox}{width=\columnwidth}
        \begin{tabular}{lccccccccc}
            \toprule
             Model & Method & AP$^{50}$ & AP$^S$ & AP$^M$ & AP$^L$ & Recall$^{\bm{C} }$ &  Recall$^{PC}$ & Recall$^{NC}$ & Precision  \\
            \midrule
             & ${FL_{\bm{\kappa}}}$ &  0.440 & 0.356 & 0.729 & 0.689 & 0.906 & 0.803 & 0.492 & \textbf{0.778}\\ 
            RetinaNet$-$50 & ${FL_{\bm{\kappa_c}}}$ & \textbf{0.445} & \textbf{0.358} & \textbf{0.741} & 0.689 & \textbf{0.918} & \textbf{0.810} & \textbf{0.505} & 0.741\\
             & ${{FL}_{\bm{\kappa_d}}}$ & 0.441 & 0.353 & 0.724 & \textbf{0.796} & 0.906 & 0.800 & 0.503 & 0.721\\
            \hline
              & ${FL_{{\bm{\kappa}}}}$ & \textbf{0.474} & \textbf{0.422} & 0.663 & 0.594 & \textbf{0.950} & 0.879 & 0.669 & \textbf{0.322}\\ 
             FCOS$-$50 &${FL_{\bm{\kappa_c}}}$ & 0.467 & 0.418 & \textbf{0.690} & 0.617 & 0.925 & 0.881 & \textbf{0.670} & 0.315\\
              & ${FL_{\bm{\kappa_d}}}$ & 0.463 & 0.416 & 0.681 & \textbf{0.637} & \textbf{0.950} & \textbf{0.888} & 0.669 & 0.291\\
            \bottomrule
        \end{tabular}
    \end{adjustbox}
\end{table}


\section{Conclusion and Future Work}
This work presents a novel safety-adapted focal loss that leverages the per-pedestrian criticality during the training to mitigate the occurrence of critical misdetections, \ie~false negatives.
We evaluate the novel loss on the safety-critical zone defined by a TTC$_{crit}<1.7\,s$ and d$_{crit}<20\,m$ and show that for RetinaNet-50 and FCOS-50 we achieve a recall increase of 2.5$\%$ and 3.2$\%$, respectively.
Supplementary, we demonstrate that the novel loss maintains stable overall performance for pedestrians outside the safety-critical zone.
This, in particular, enables the employment of the safety-adapted focal loss for AD applications as the initial concept provides promising results. 

Up to now, we have only considered the mitigation of false negatives, but from a safety perspective, false positives are also of great concern. An approach to determine the criticality of falsely detected pedestrians should be included in future work.
Further, we plan to extend the safety-adapted loss to the regression task to improve the detection quality of critical pedestrians.

\bibliographystyle{IEEEtran}
\bibliography{bibliography}

\begin{thebibliography}{10}
\providecommand{\url}[1]{#1}
\csname url@samestyle\endcsname
\providecommand{\newblock}{\relax}
\providecommand{\bibinfo}[2]{#2}
\providecommand{\BIBentrySTDinterwordspacing}{\spaceskip=0pt\relax}
\providecommand{\BIBentryALTinterwordstretchfactor}{4}
\providecommand{\BIBentryALTinterwordspacing}{\spaceskip=\fontdimen2\font plus
\BIBentryALTinterwordstretchfactor\fontdimen3\font minus
  \fontdimen4\font\relax}
\providecommand{\BIBforeignlanguage}[2]{{%
\expandafter\ifx\csname l@#1\endcsname\relax
\typeout{** WARNING: IEEEtran.bst: No hyphenation pattern has been}%
\typeout{** loaded for the language `#1'. Using the pattern for}%
\typeout{** the default language instead.}%
\else
\language=\csname l@#1\endcsname
\fi
#2}}
\providecommand{\BIBdecl}{\relax}
\BIBdecl

\bibitem{FCOS2019}
Z.~Tian, C.~Shen, H.~Chen, and T.~He, ``Fcos: Fully convolutional one-stage
  object detection,'' 2019.

\bibitem{web}
\BIBentryALTinterwordspacing
``viso.ai,'' 2023, accessed on September 14, 2023. [Online]. Available:
  \url{https://viso.ai/applications/computer-vision-applications/}
\BIBentrySTDinterwordspacing

\bibitem{philion2020learning}
J.~Philion, A.~Kar, and S.~Fidler, ``Learning to evaluate perception models
  using planner-centric metrics,'' 2020.

\bibitem{Brkle2021SafePO}
C.~B{\"u}rkle, F.~Oboril, J.~Jarquin, F.~Pasch, and K.-U. Scholl, ``Safe
  perception: On relevance of objects for vehicle safety,'' \emph{2021 IEEE
  International Intelligent Transportation Systems Conference (ITSC)}, pp.
  3957--3964, 2021.

\bibitem{Mori2023ConservativeEO}
\BIBentryALTinterwordspacing
K.~Mori, K.~Storms, and S.~C. Peters, ``Conservative estimation of perception
  relevance of dynamic objects for safe trajectories in automotive scenarios,''
  \emph{ArXiv}, vol. abs/2307.10873, 2023. [Online]. Available:
  \url{https://api.semanticscholar.org/CorpusID:259991178}
\BIBentrySTDinterwordspacing

\bibitem{Cheng2020SafetyAwareHO}
C.-H. Cheng, ``Safety-aware hardening of 3d object detection neural network
  systems,'' in \emph{International Conference on Computer Safety, Reliability,
  and Security}, 2020.

\bibitem{Willers2020SafetyCA}
O.~Willers, S.~Sudholt, S.~Raafatnia, and S.~Abrecht, ``Safety concerns and
  mitigation approaches regarding the use of deep learning in safety-critical
  perception tasks,'' in \emph{SAFECOMP Workshops}, 2020.

\bibitem{DAmour2020UnderspecificationPC}
A.~D'Amour, K.~A. Heller, D.~I. Moldovan, B.~Adlam, B.~Alipanahi, A.~Beutel,
  C.~Chen, J.~Deaton, J.~Eisenstein, M.~D. Hoffman, F.~Hormozdiari, N.~Houlsby,
  S.~Hou, G.~Jerfel, A.~Karthikesalingam, M.~Lucic, Y.-A. Ma, C.~Y. McLean,
  D.~Mincu, A.~Mitani, A.~Montanari, Z.~Nado, V.~Natarajan, C.~Nielson, T.~F.
  Osborne, R.~Raman, K.~Ramasamy, R.~Sayres, J.~Schrouff, M.~G. Seneviratne,
  S.~Sequeira, H.~Suresh, V.~Veitch, M.~Vladymyrov, X.~Wang, K.~Webster,
  S.~Yadlowsky, T.~Yun, X.~Zhai, and D.~Sculley, ``Underspecification presents
  challenges for credibility in modern machine learning,'' \emph{J. Mach.
  Learn. Res.}, vol.~23, pp. 226:1--226:61, 2020.

\bibitem{wolf}
M.~Wolf, L.~R. Douat, and M.~Erz, ``Safety-aware metric for people detection,''
  in \emph{2021 IEEE International Intelligent Transportation Systems
  Conference (ITSC)}, 2021, pp. 2759--2765.

\bibitem{bansal2021risk}
A.~Bansal, J.~Singh, M.~Verucchi, M.~Caccamo, and L.~Sha, ``Risk ranked recall:
  Collision safety metric for object detection systems in autonomous
  vehicles,'' in \emph{2021 10th Mediterranean Conference on Embedded Computing
  (MECO)}.\hskip 1em plus 0.5em minus 0.4em\relax IEEE, 2021, pp. 1--4.

\bibitem{LGH+22}
M.~Lyssenko, C.~Gladisch, C.~Heinzemann, M.~Woehrle, and R.~Triebel, ``{Towards
  Safety-Aware Pedestrian Detection in Autonomous Systems},'' in \emph{Proc. of
  IROS}, Kyoto, Japan, Oct. 2022, pp. 293--300.

\bibitem{nuscenes2019}
H.~Caesar, V.~Bankiti, A.~H. Lang, S.~Vora, V.~E. Liong, Q.~Xu, A.~Krishnan,
  Y.~Pan, G.~Baldan, and O.~Beijbom, ``nuscenes: A multimodal dataset for
  autonomous driving,'' \emph{arXiv preprint arXiv:1903.11027}, 2019.

\bibitem{retina}
T.-Y. Lin, P.~Goyal, R.~Girshick, K.~He, and P.~Dollár, ``Focal loss for dense
  object detection,'' in \emph{2017 IEEE International Conference on Computer
  Vision (ICCV)}, 2017, pp. 2999--3007.

\bibitem{Zaidi2021ASO}
S.~S.~A. Zaidi, M.~S. Ansari, A.~Aslam, N.~Kanwal, M.~N. Asghar, and B.~Lee,
  ``A survey of modern deep learning based object detection models,''
  \emph{Digit. Signal Process.}, vol. 126, p. 103514, 2021.

\bibitem{Andrea2022}
A.~Ceccarelli and L.~Montecchi, ``Evaluating the consequences of object
  (mis)detection from a safety and reliability perspective: Discussion and
  measures,'' 2022.

\bibitem{LGH+21}
M.~Lyssenko, C.~Gladisch, C.~Heinzemann, M.~Woehrle, and R.~Triebel, ``From
  evaluation to verification: Towards task-oriented relevance metrics for
  pedestrian detection in safety-critical domains,'' in \emph{Workshop on Safe
  Artificial Intelligence for Automated Driving}, 2021.

\bibitem{Gannamaneni2021SemanticCT}
S.~S. Gannamaneni, S.~Houben, and M.~Akila, ``Semantic concept testing in
  autonomous driving by extraction of object-level annotations from carla,''
  \emph{2021 IEEE/CVF International Conference on Computer Vision Workshops
  (ICCVW)}, pp. 1006--1014, 2021.

\bibitem{Interactiondynamics}
S.~Topan, K.~Leung, Y.~Chen, P.~Tupekar, E.~Schmerling, J.~Nilsson, M.~Cox, and
  M.~Pavone, ``Interaction-dynamics-aware perception zones for obstacle
  detection safety evaluation,'' in \emph{2022 IEEE Intelligent Vehicles
  Symposium (IV)}, 2022, pp. 1201--1210.

\bibitem{safetyzones}
S.~Topan, Y.~Chen, E.~Schmerling, K.~Leung, J.~Nilsson, M.~Cox, and M.~Pavone,
  ``Refining obstacle perception safety zones via maneuver-based
  decomposition,'' in \emph{2023 IEEE Intelligent Vehicles Symposium (IV)},
  2023, pp. 1--8.

\bibitem{autoped}
Y.~Tang, B.~Li, M.~Liu, B.~Chen, Y.~Wang, and W.~Ouyang, ``Autopedestrian: An
  automatic data augmentation and loss function search scheme for pedestrian
  detection,'' \emph{IEEE Transactions on Image Processing}, vol.~30, pp.
  8483--8496, 2021.

\bibitem{BoLi2022}
B.~Li, Y.~Yao, J.~Tan, G.~Zhang, F.~Yu, J.~Lu, and Y.~Luo, ``Equalized focal
  loss for dense long-tailed object detection,'' 2022.

\bibitem{Liao2023}
H.-C. Liao, C.-H. Cheng, H.~Esen, and A.~Knoll, ``Improving the safety of 3d
  object detectors in autonomous driving using iogt and distance measures,''
  2023.

\bibitem{Schneider2021}
P.~Schneider, M.~Butz, C.~Heinzemann, J.~Oehlerking, and M.~Woehrle, ``Towards
  threat metric evaluation in complex urban scenarios,'' in \emph{2021 IEEE
  International Intelligent Transportation Systems Conference (ITSC)}, 2021,
  pp. 1192--1198.

\bibitem{6784493}
M.~Althoff and J.~M. Dolan, ``Online verification of automated road vehicles
  using reachability analysis,'' \emph{IEEE Transactions on Robotics}, vol.~30,
  no.~4, pp. 903--918, 2014.

\bibitem{DROF19}
J.~{Dahl}, G.~R. de~Campos, C.~Olsson, and J.~Fredriksson, ``Collision
  avoidance: A literature review on threat-assessment techniques,'' \emph{IEEE
  Transactions on Intelligent Vehicles}, vol.~4, no.~1, pp. 101--113, March
  2019.

\bibitem{binaryTTC}
A.~Badki, O.~Gallo, J.~Kautz, and P.~Sen, ``Binary {TTC:} {A} temporal geofence
  for autonomous navigation,'' \emph{CoRR}, vol. abs/2101.04777, 2021.

\bibitem{nuscapi}
``nuscenes-devkit,'' \url{https://github.com/nutonomy/nuscenes-devkit}, 2021.

\bibitem{nuscerr}
\BIBentryALTinterwordspacing
``Possible issues with annotations,'' 2022, accessed on September 14, 2023.
  [Online]. Available:
  \url{https://github.com/nutonomy/nuscenes-devkit/issues/366}
\BIBentrySTDinterwordspacing

\bibitem{pytorch}
\BIBentryALTinterwordspacing
A.~Paszke, S.~Gross, F.~Massa, A.~Lerer, J.~Bradbury, G.~Chanan, T.~Killeen,
  Z.~Lin, N.~Gimelshein, L.~Antiga, A.~Desmaison, A.~K{\"o}pf, E.~Yang,
  Z.~DeVito, M.~Raison, A.~Tejani, S.~Chilamkurthy, B.~Steiner, L.~Fang,
  J.~Bai, and S.~Chintala, ``Pytorch: An imperative style, high-performance
  deep learning library,'' in \emph{Neural Information Processing Systems},
  2019. [Online]. Available:
  \url{https://api.semanticscholar.org/CorpusID:202786778}
\BIBentrySTDinterwordspacing

\bibitem{retinayeh}
Y.~Henon, ``pytorch-retinanet,''
  \url{https://github.com/yhenon/pytorch-retinanet}, 2021.

\bibitem{fcostian}
Z.~Tian, ``Fcos,'' \url{https://github.com/tianzhi0549/FCOS}, 2021.

\bibitem{ZW23}
X.~Zhou and O.~Wu, ``Which samples should be learned first: Easy or hard?''
  \emph{IEEE Transactions on Neural Networks and Learning Systems}, pp. 1--15,
  2023.

\end{thebibliography}
\end{document}